\documentclass[11pt]{article}

\usepackage[preprint]{acl}

\usepackage{times}
\usepackage{latexsym}

\usepackage[T1]{fontenc}

\usepackage[utf8]{inputenc}

\usepackage{microtype}

\usepackage{inconsolata}

\usepackage{graphicx}

\usepackage{hyperref}       
\usepackage{url}            
\usepackage{booktabs}       
\usepackage{amsfonts}       
\usepackage{nicefrac}       
\usepackage{microtype}      
\usepackage{xcolor}         
\usepackage{soul}
\usepackage{graphicx}
\usepackage{amsmath}
\usepackage{multirow}
\usepackage{tabularx}
\usepackage{colortbl}
\usepackage{wrapfig} 
\usepackage{pifont}
\usepackage{subcaption}

\title{GMSA: Enhancing Context Compression via Group Merging and Layer Semantic Alignment}

\author{
 \textbf{Jiwei Tang\textsuperscript{1}}\thanks{\ \ indicates equal contribution.},
 \textbf{Zhicheng Zhang\textsuperscript{1*}},
 \textbf{Shunlong Wu\textsuperscript{1}},
 \textbf{Jingheng Ye\textsuperscript{1}},
 \textbf{Lichen Bai\textsuperscript{1}},
\\
 \textbf{Zitai Wang\textsuperscript{1}},
 \textbf{Tingwei Lu\textsuperscript{1}},
 \textbf{Lin Hai\textsuperscript{1}},
 \textbf{Yiming Zhao\textsuperscript{4}},
 \textbf{Hai-Tao Zheng\textsuperscript{1,2}\thanks{Corresponding authors.}},
 \textbf{Hong-Gee Kim\textsuperscript{3}}
\\
 \textsuperscript{1}Tsinghua University \hspace{0.6mm} \textsuperscript{2}Pengcheng Laboratory \\
 \textsuperscript{3}Seoul National University \hspace{0.6mm} \textsuperscript{4}Sun Yat-sen University
\\
 \texttt{tangjw24@mails.tsinghua.edu.cn} \\
 \\
}

\begin{document}
\maketitle


\begin{abstract}
Large Language Models (LLMs) have achieved remarkable performance across a wide range of Natural Language Processing (NLP) tasks. However, in long-context scenarios, they face two challenges: high computational cost and information redundancy. To address these challenges, we propose \textbf{GMSA}, an encoder--decoder context compression framework that generates a compact sequence of soft tokens for downstream tasks. GMSA introduces \textbf{Group Merging} to achieve more uniform aggregation, mitigating semantic dominance during autoencoder pretraining, and \textbf{Layer Semantic Alignment (LSA)} to bridge the semantic gap between high-level abstract semantics and low-level input semantics. We first pretrain GMSA as an autoencoder and then fine-tune it for downstream tasks. Experiments demonstrate that GMSA improves context reconstruction compared to existing soft prompt compression paradigm and outperforms baselines on multiple long-context question answering and summarization benchmarks across two backbone models, while maintaining low end-to-end latency.
\end{abstract}

\section{Introduction}
Thanks to powerful reasoning and generalization capabilities, Large Language Models (LLMs) have achieved remarkable performance across various Natural Language Processing (NLP) tasks~\cite{qwen2025qwen25technicalreport,team2025kimi,liu2025deepseek,zeng2025glm,zhao2025cosoptimaleventscheduling}. However, directly applying LLMs to long-context scenarios presents two challenges: (1) Computational inefficiency. When processing long prompts, the quadratic complexity of the Transformer's attention mechanism~\cite{vaswani2017attention} results in computational inefficiency. (2) Redundant information. Much redundant information in long-context scenarios can degrade model performance~\cite{tang-etal-2025-perception}.

Prompt compression methods address these two challenges by significantly reducing input length and redundant information. It can be categorized into hard prompt compression~\cite{li-etal-2023-compressing,jiang-etal-2023-llmlingua,pan-etal-2024-llmlingua,jiang-etal-2024-longllmlingua,tang-etal-2025-perception,zhou2025mooscomp,cao2025efpcefficientflexibleprompt,chen2025pis,Zhao_Wu_Xu_2025} and soft prompt compression~\cite{mu2023learning,ge2024incontext,li2025500xcompressor, liao2025hardsofthybridcontext,dai-etal-2025-pretraining, 10.1145/3701551.3703527,DBLP:conf/iclr/Zhang0XSYD25,DBLP:journals/corr/abs-2509-15763,tang2026readhumancompressingcontext,tang2026comicoarsetofinecontextcompression,lv2026datadistributionmattersdatacentric,zhao2026cometcollaborativememorytransformer}. Hard prompt compression methods achieve compression by deleting certain tokens from the original context or generating a summary. However, such explicit compression inevitably compromises semantic integrity. In contrast, by leveraging the inherent redundancy in semantic vectors~\cite{ethayarajh-2019-contextual,aghajanyan-etal-2021-intrinsic}, soft prompt compression learns a set of soft tokens that is much shorter than the original context while preserving more complete semantic information.  
\begin{figure*}[htb]
    \centering
    \includegraphics[width=0.95\linewidth]{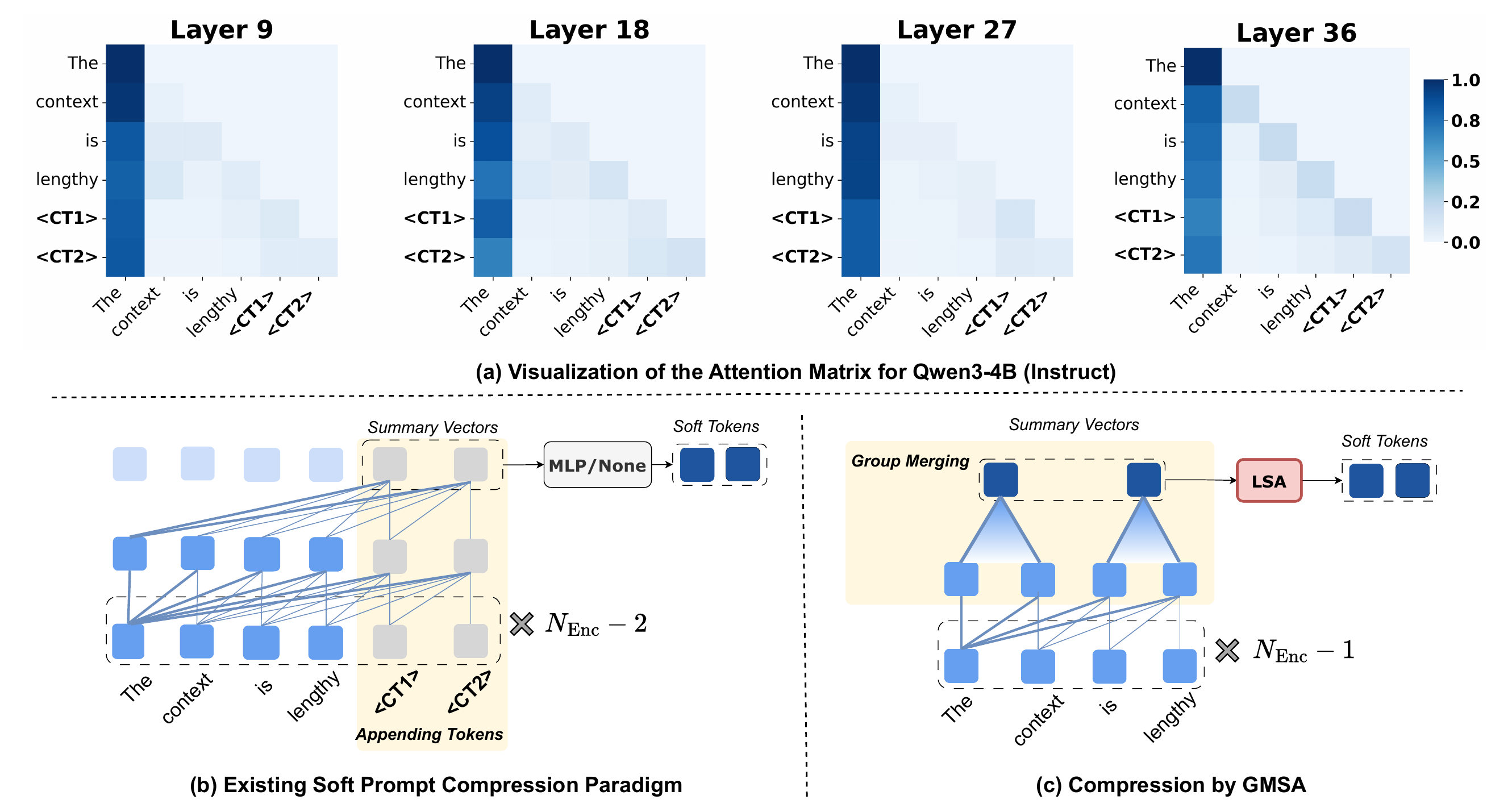}
    \caption{Existing Soft Prompt Compression Paradigm vs. Compression by GMSA. (a) visualizes the attention matrix when processing ``\textit{The context is lengthy <CT1> <CT2>}'', where \textit{<CT1>} and \textit{<CT2>} are randomly initialized tokens. (b) represents existing soft prompt compression methods. It first learns summary vectors layer by layer, and then either directly use the summary vectors as soft tokens (marked as “None” in (b)) or obtain soft tokens via a MultiLayer Perceptron (MLP), where $N_{\text{Enc}}$ denotes the number of encoder layers. (c) denotes the compression paradigm of GMSA, which first learns summary vectors through group merging and completes semantic alignment via the Layer Semantic Alignment (LSA) module.}
    \label{fig:intro}
\end{figure*}

However, existing soft prompt compression methods have two limitations: (1) Semantic dominance in autoencoder pretraining process. LLM tends to aggregate information on a few anchor tokens~\cite{NEURIPS2023_6ceefa7b,xiao2023streamingllm,huang2024opera,DBLP:journals/corr/abs-2505-06708}. As shown in Figure~\ref{fig:intro}, in existing soft prompt compression paradigm, the appending randomly initialized tokens (``\textit{<CT1>}'' and ``\textit{<CT2>}'') learn summary vectors layer by layer. The semantics of anchor token (``\textit{The}'') is emphasized layer by layer, resulting in the semantics of the summary vectors being dominated by it while the semantics of other tokens (``\textit{context}'', ``\textit{is}'', and ``\textit{lengthy}'') are diluted. This limits the retention of \emph{complete semantics} in autoencoder pretraining, which is a common approach in soft prompt compression~\cite{ge2024incontext,cheng2024xrag, li2025500xcompressor,liao2025hardsofthybridcontext,dai-etal-2025-pretraining,10.1145/3701551.3703527}. (2) Ignoring the large semantic gap between different layers of the LLMs~\cite{liu-etal-2024-fantastic,jin-etal-2025-exploring}, which cannot be directly bridged by the MLP layer (as demonstrated by the \emph{w/o} LSA results in Tab.~\ref{tab:ablation} and Fig.~\ref{fig:loss_comparison}). The summary vectors, which represent high-level abstract semantics, are directly treated as ordinary tokens (i.e., as low-level semantics) and directly fed into the decoder during training and testing, resulting in a large semantic gap. Therefore, two research questions naturally arise: (1) \textit{How can we mitigate semantic dominance in soft prompt compression pretraining process?} (2) \textit{How can we bridge the large semantic gap between different layers?}

To this end, we propose GMSA, a context compression framework based on the encoder-decoder architecture that addresses these limitations via \textbf{Group Merging} and \textbf{Layer Semantic Alignment (LSA)}. Specifically, group merging partitions the input tokens into equal-sized groups and compresses each group via average pooling. \emph{This uniform aggregation mitigates the dominance of anchor tokens, therefore helping to preserve complete semantics during pretraining.} Furthermore, we introduce a Layer Semantic Alignment (LSA) module to bridge the semantic gap between high-level abstract summary vectors and low-level input semantics. LSA is implemented as a small stack of Transformer blocks, initialized with the weights of the lower decoder layers (Figure~\ref{fig:framework}), thereby inheriting the representation space of low-level semantics. \emph{By feeding the summary vectors through LSA, we project them from a high-level semantic space into the lower-level space, which alleviates cross-layer semantic gap.} We first pretrain GMSA as an autoencoder to encourage the generated soft tokens to retain complete semantics, and then fine-tune it on downstream tasks.

Our contributions are threefold: (1) We identify and analyze two limitations in existing soft prompt compression paradigm: (i) semantic dominance in soft prompt compression pretraining process, and (ii) a semantic gap arising from the direct use of high-level summary vectors as decoder inputs. (2) We propose GMSA, a context compression framework that introduces (i) Group Merging to mitigate semantic dominance in soft compression pretraining process, and (ii) a Layer Semantic Alignment (LSA) module bridges semantic gap. (3) We conduct extensive experiments on context reconstruction, diverse benchmarks, demonstrating that GMSA achieves high semantic fidelity and superior downstream performance compared to baselines, while incurring low end-to-end latency.


\section{Problem Formulation}
Given a retrieval-augmented prompt $X = (X^{\text{ins}}, X^{d_{1}}, ..., X^{d_{k}}, ..., X^{d_{K}}, X^{\text{q}})$, where $X^{ins}$, $\{X^{d_{k}}\}^{K}_{k=1}$, and $X^{\text{q}}$ represent the instruction, context, and input question respectively. The prompt has a total token length $L$. The key aspect of the context compression system lies in generating a compressed prompt $\widetilde{X}$ with length $\widetilde{L}$, where the compression rate is defined as $\tau=\frac{L}{\widetilde{L}}$. Let $y$ denote the ground truth answer given the original input $X$, and $\widetilde{y}$ denote the answer generated by the large language model (LLM) when input with the compressed prompt $\widetilde{x}$. We aim for the distributions of $y$ and $\widetilde{y}$ to be similar under high compression rates $\tau$. This can be formulated as:
\begin{equation}
\min _{\widetilde{\boldsymbol{x}}, \tau} \operatorname{KL}\left(P\left(\widetilde{y} \mid \widetilde{X}\right), P\left(y \mid X \right)\right) \, .
\end{equation}

\section{Related Work}
\label{apx:rela_work}

\paragraph{Hard Prompt Compression.} Hard prompt compression refers to the removal of some less important tokens from the original prompt or the generation of summaries to achieve compression. The compressed prompt is explicit text. It can mainly be divided into the following four categories: (1) Perplexity-based methods. Selective-Context~\cite{li-etal-2023-compressing} removes certain lexical units based on perplexity, while methods such as LLMLingua~\cite{jiang-etal-2023-llmlingua}, LongLLMLingua~\cite{jiang-etal-2024-longllmlingua}, and Perception Compressor~\cite{tang-etal-2025-perception} adopt a coarse-to-fine framework to gradually eliminate less important parts. (2) Bidirectional semantic-based methods. Considering the unidirectional nature of perplexity, some approaches employ bidirectional semantic information for compression, such as LLMLingua-2~\cite{pan-etal-2024-llmlingua}, MOOSComp~\cite{zhou2025mooscomp}, and EFPC~\cite{cao2025efpcefficientflexibleprompt}. (3) Methods based on intrinsic attention mechanisms. Compression is achieved through the intrinsic attention mechanisms of LLMs, such as PIS~\cite{chen2025pis} and AttnComp~\cite{Zhao_Wu_Xu_2025}. (4) Summary generation. This involves generating linguistic summaries that contain useful information for long text content, such as CompACT~\cite{yoon-etal-2024-compact} and RECOMP~\cite{xu2024recomp}. \textit{Although these methods improve the computational efficiency of inference through prompt compression, they compromise the semantic integrity of the original prompt.}


\paragraph{Soft Prompt Compression.} Soft prompt compression has become a research hotspot in the field of Natural Language Processing (NLP). The goal of soft prompt compression is to learn a set of soft tokens (with a sequence length much shorter than the original text) to achieve compression, where the compressed soft prompts cannot be explicitly converted into text. Among existing methods, xRAG~\cite{cheng2024xrag} focuses on processing short texts and extreme compression. More mainstream methods, such as GIST~\cite{mu2023learning}, AutoCompressor~\cite{chevalier2023adapting}, 500xCompressor~\cite{li2024500xcompressorgeneralizedpromptcompression}, ICAE~\cite{ge2024incontext} and VoCo-LLaMA~\cite{ye2024voco}, learn soft tokens in an autoregressive manner by appending randomly initialized additional tokens. This leads to the semantics of anchor tokens in the input sequence being increasingly emphasized layer by layer, while the semantics of other tokens are diluted and cannot be fully preserved in the summary vectors. Moreover, these methods only use Multilayer Perceptrons (MLPs) for coarse-grained semantic alignment when semantic alignment is required, ignoring the significant differences in representations across different layers of large models. \textit{Our proposed method mitigates semantic dominance during autoencoder pretraining via group merging, which uniformly aggregates within each equal-sized group. Moreover, it bridges the semantic gap across LLM layers through a Layer Semantic Alignment (LSA) module that inherits the representation space of low-level semantics.}

\section{GMSA}

\begin{figure*}[tbh]
    \centering
    \includegraphics[width=1\linewidth]{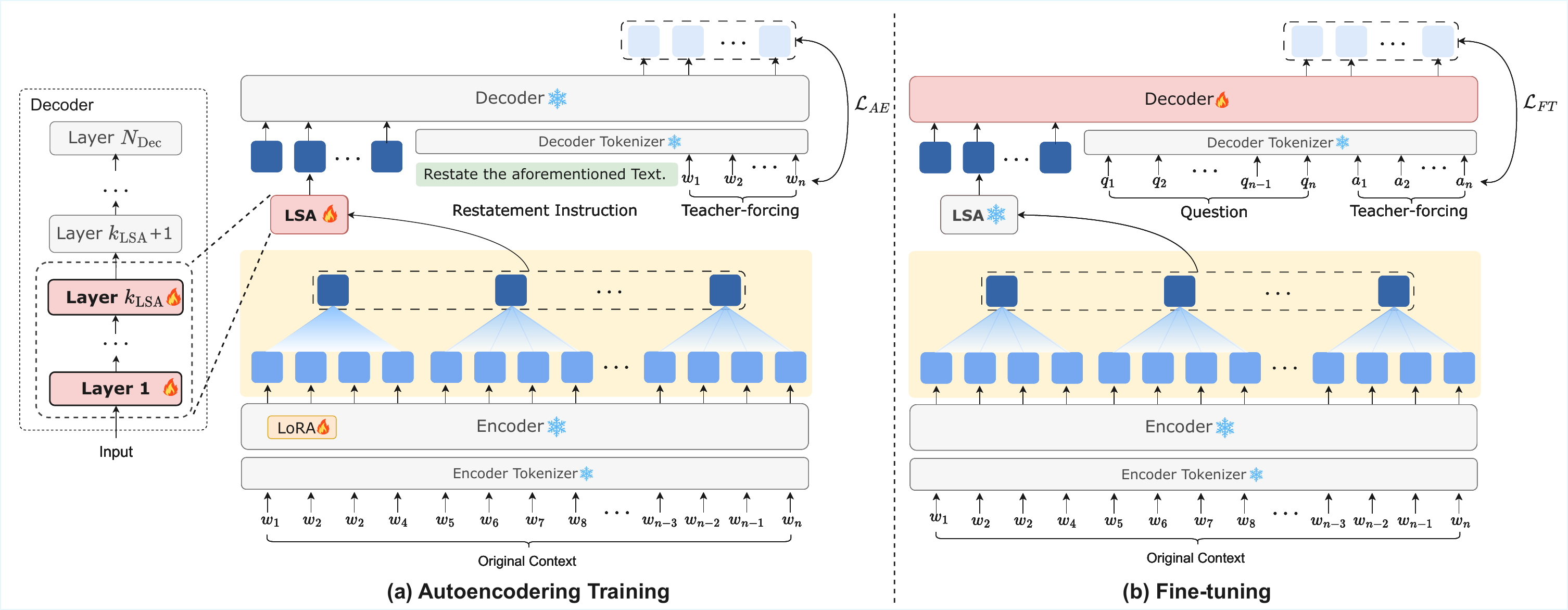}
    \caption{\textbf{The Training process of GMSA.} GMSA consists of an encoder and a decoder and is trained in two stages: (a) autoencoding pretraining, followed by (b) task-specific fine-tuning. During autoencoding training, only the encoder and the Layer Semantic Alignment (LSA) module are trained to reconstruct the original context, enabling GMSA to produce semantically complete compressed representations. During fine-tuning, only the decoder is trained, and GMSA is optimized end-to-end on question answering format to enhance its ability to extract knowledge from the compressed representations and improve downstream performance. Notably, a single LSA layer is sufficient to preserve semantics effectively (see Appendix~\ref{apx:lsa_lays}), so $k_{\text{LSA}} \ll N_{\text{Dec}}$.}
    \label{fig:framework}
\end{figure*}

In this section, we elaborate on the architecture of our proposed context compression framework, GMSA, which includes two key components: group merging and layer semantic alignment (LSA). GMSA undergoes a two-stage training process: autoencoder pretraining and fine-tuning (see Figure~\ref{fig:framework}). First, GMSA ensures that the generated soft tokens contain the complete semantic representation of the original text through the autoencoder pretraining process. Then, it applies the knowledge contained in the soft tokens to downstream tasks via fine-tuning.


\subsection{Group Merging}
\paragraph{Extraction of Semantic Features.}
We extract the semantic features of the original text through a language model (e.g., Qwen3-4B) as the encoder. The encoder is trained using LoRA.
\begin{equation}
    H = \texttt{Encoder}(X),
\end{equation}
where $X$ is the original text and $H$ is the obtained last hidden state.

\paragraph{Merging.}
We divide the obtained $H$ into several groups according to the size of the compression limit, as the group length $L_G$ (e.g., when the compression rate is 4, the group length is also 4). To this end, original text representations are organized as follows:
\begin{equation}
\begin{split}
H & =\left[H_{\mathbf{}_{1}}, \ldots, H_{\mathbf{G}_{N_{g}}}\right] \\
  & =\left[H_{1: L_G}, \ldots, H_{N_{d}-L_G+1: N_{d}}\right].
\end{split}
\end{equation}
We take the average of each dimension of each group token to obtain the initial compressed representation.
\begin{equation}
\begin{split}
    \widetilde{H} & = \left[ \bar{H}_{\mathbf{G}_{1}},\ldots,\bar{H}_{\mathbf{G}_{N}} \right] \\
    & = \left[ \frac{1}{L_G}\sum H_{\mathbf{G}_{1}}, \ldots,\frac{1}{L_G}\sum H_{\mathbf{G}_{N}}  \right] \, ,
\end{split}
\end{equation}
where $\widetilde{H}$ is the obtained initial compressed representation.

\subsection{Layer Semantic Alignment}
The layer semantic alignment (LSA) module is used to complete the alignment from the soft tokens generated by the encoder (high-level semantics) to the primary semantics of the decoder. Given the significant differences in semantic representation between different layers of large language models (LLMs), the LSA is trained via full fine-tuning.
\begin{equation}
    \widetilde{m} = \mathcal{F}_{k_{\text{LSA}}}(\widetilde{H}),
\end{equation}
where $H$ is the final compressed representation, $\mathcal{F}_{k_{\text{LSA}}}$ denotes Transformer blocks initialized with the weights from the first $k$ layers of the decoder, and $\widetilde{m}$ denotes the generated soft tokens. Just one layer of LSA is sufficient to achieve excellent semantic preservation (for space limitations, please refer to Appendix~\ref{apx:lsa_lays}), so in this work, we can just set $k_{\text{LSA}}=1$.



\subsection{Autoencoder Pretraining}
The Autoencoder Pretraining process, which aims to encode the complete information of the original text into memory embeddings, is achieved through autoencoder-based training. We hope to minimize the loss of the reconstructed text, which can be expressed as:
\begin{equation}
    \mathcal{L}_{AE}=-\sum_{i=1} \log p_{\phi}\left(x_{i} \mid \widetilde{m}, X^{\text {ins}}, x_{<i}\right),
\end{equation}
where $p_{\phi}(\cdot)$ is the decoder probability distribution obtained after the softmax function, and $x_{i}$ is the $i$-th token in the original text.

\subsection{Fine-tuning}
After completing autoencoder pretraining, we need to teach the decoder how to utilize the soft tokens. We achieve this by performing full fine-tuning of the decoder, which can be expressed as:
\begin{equation}
    \mathcal{L}_{\mathrm{FT}}=-\sum_{i=1}^{n} \log p_{\phi}\left(a_{i} \mid \widetilde{m}, q_1,...,q_n,a_{<i}\right),
\end{equation}
where $p_{\phi}(\cdot)$ is the decoder probability distribution obtained after the softmax function, and $a_i$ denotes the $i$-th token in the predicted answer. 


\section{Experiments}
In this section, we attempt to answer the following research questions (RQs): (1) How effective is GMSA in context reconstruction (RQ1)? (2) How does GMSA utilize knowledge compared with other baselines (RQ2)? (3) How effective are the individual components of GMSA (RQ3)? 
\subsection{Settings}

\paragraph{Training.} GMSA involves a two-stage training process: autoencoder pretraining and fine-tuning. We use  datasets: PwC~\cite{ge2024incontext}, NaturalQuestions~\cite{liu-etal-2024-lost}, 2WikiMQA~\cite{ho-etal-2020-constructing}, HotpotQA~\cite{yang-etal-2018-hotpotqa}, NarrativeQA~\cite{DBLP:journals/tacl/KociskySBDHMG18}, MultiNews~\cite{DBLP:conf/acl/FabbriLSLR19} (see more details on Appendix~\ref{apx:data_details}). Among them, we use PwC to evaluate the performance of context reconstruction, while the other datasets are employed to measure downstream knowledge application. During training, we randomly sample compression rates (i.e., 4x compression and 8x compression) for each training sample. We set the batch size to 32 and trained all models for one epoch. The learning rate is set to $1 \times 10^{-4}$ during the autoencoder pretraining stage and to $1 \times 10^{-5}$ during the fine-tuning stage. We train GMSA under two distinct experimental configurations: (1) We perform autoencoder pretraining on the PwC dataset to get \textbf{GMSA-AE}. (2) We develop the general-purpose \textbf{GMSA} for all remaining tasks. This is achieved through total training process: autoencoder pretraining followed by fine-tuning on a balanced hybrid dataset. This dataset is constructed by sampling 20,000 instances from each of the following sources: NaturalQuestions, 2WikiMQA, HotpotQA, NarrativeQA, and MultiNews.

\paragraph{Implementation.} GMSA is implemented based on LLaMA-3.2-3B (Instruct) and Qwen3-4B (Instruct). Due to the GPU memory constraints of our computational resources, the maximum input length is set to 12K tokens for autoencoding training and 32K tokens for fine-tuning. To ensure fair comparison, all baseline results are obtained from our re-implementations based on official open-source code. All experiments are conducted on 8 NVIDIA H20 GPUs.

\paragraph{Evaluation Metrics.}
For the context reconstruction task on the PwC dataset, we use BLEU~\cite{papineni2002bleu}, Prefix Exact Match, BERT Score~\cite{Zhang*2020BERTScore:}, and ROUGE~\cite{lin2004rouge} for evaluation. For the QA tasks on Natural Questions, TriviaQA, and 2WikiMQA, we use Exact Match (EM)~\cite{lewis2020retrieval} and F1~\cite{yang-etal-2018-hotpotqa} score for evaluation. For Summary task on MultiNews, we use F1 score too.

\paragraph{Baselines.} For the task of context reconstruction, we train a \textbf{I}n\textbf{C}ontext \textbf{A}uto\textbf{E}ncoder (i.e., ICAE-AE) as a baseline \emph{only} using autoencoder pretraining~\cite{ge2024incontext} and the same training hyperparameters as GMSA. In terms of downstream knowledge application, we conduct comprehensive comparisons with various methods, including: Hard prompt compression (e.g., LongLLMLingua~\cite{jiang-etal-2024-longllmlingua}, LLMLingua-2-large~\cite{pan-etal-2024-llmlingua}, Provence~\cite{chirkova2025provenceefficientrobustcontext}, EXIT~\cite{hwang2025exitcontextawareextractivecompression}), and soft prompt compression (e.g., ICAE~\cite{ge2024incontext}, Activation Beacon~\cite{zhang2024long}). Additionally, we compare against the original input prompt and zero-shot prompting to establish upper and lower bounds on original model performance.


\subsection{Main Result}

\begin{figure*}
    \centering
    \includegraphics[width=1\linewidth]{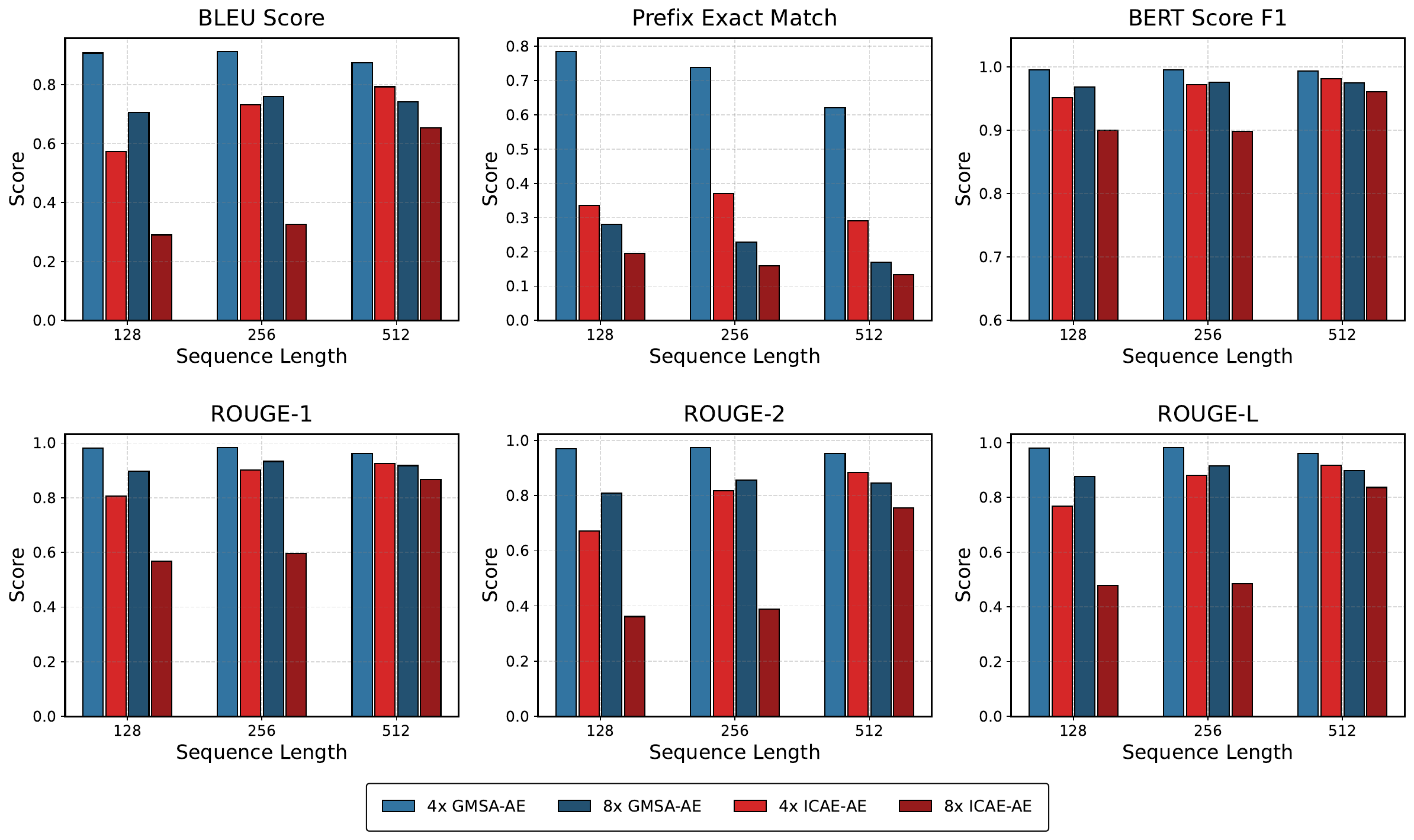}
    \caption{GMSA-AE vs. ICAE-AE on the context reconstruction task (PwC dataset). Sequence Length represents different context reconstruction lengths (i.e., 128, 256, 512).}
    \label{fig:restatement_comparasion}
\end{figure*}
\begin{table*}[htb]
\centering
\fontsize{8}{9.1}\selectfont
\caption{Main results on benchmarks. We \textbf{bold} the optimal results. Closed-book indicates using only the input question as the input, while Original Prompt indicates using original context as the input. }
\label{tab:overall_comparison}
\begin{tabular}{l | c c c c c c c c c | c c}
\toprule
\multirow{2}{*}{\textbf{Methods}} & 
  \multicolumn{2}{c}{\textbf{NaturalQA}} & 
  \multicolumn{2}{c}{\textbf{2WikiMQA}} & 
  \multicolumn{2}{c}{\textbf{HotpotQA}} & 
  \multicolumn{2}{c}{\textbf{NarrativeQA}} &
  \textbf{MultiNews} &
  \multicolumn{2}{c}{\textbf{AVG}} \\
\cmidrule(lr){2-3} \cmidrule(lr){4-5} \cmidrule(lr){6-7} \cmidrule(lr){8-9} \cmidrule(lr){10-10} \cmidrule(lr){11-12}
& \textbf{EM} & \textbf{F1} & \textbf{EM} & \textbf{F1} & \textbf{EM} & \textbf{F1} & \textbf{EM} & \textbf{F1} & \textbf{F1} & \textbf{EM} & \textbf{F1} \\
\midrule
\multicolumn{12}{c}{\textbf{LLaMA-3.2-3B-Instruct}} \\
\midrule
Closed-book & 7.38 & 14.01 & 2.25 & 15.37 & 4.91 & 14.68 & 0.00 & 8.34 & - & 3.64 & 13.10 \\
Original Prompt & 37.55 & 48.00 & 16.29 & 32.34 & 34.94 & 51.35 & 12.78 & 29.27 & 25.25 & 25.39 & 37.24 \\
\midrule
\multicolumn{12}{c}{\textit{4x Compression Constraint}} \\
\midrule
ICAE & 25.12 & 27.36 & 25.20 & 28.77 & 17.18 & 24.90 & 2.07 & 11.18 & 18.89 & 17.39 & 22.22 \\
LLMLingua-2-large & 25.76 & 36.55 & 15.09 & 20.00 & 26.79 & 38.22 & 9.02 & 17.90 & 28.40 & 19.17 & 28.21 \\
Activation Beacon & 34.05 & 46.53 & 33.31 & 40.92 & 37.81 & 51.30 & 10.35 & 19.09 & 27.98 & 28.88 & 37.16 \\
Provence & 32.35 & 42.28 & 30.12 & 37.46 & 38.58 & 51.24 & \textbf{10.41} & 18.64 & 28.79 & 27.87 & 35.68 \\
EXIT & 44.40 & 54.88 & 27.26 & 33.86 & 43.59 & 57.04 & 8.27 & 18.94 & 33.12 & 30.88 & 39.57 \\
LongLLMLingua & 44.97 & 55.01 & 21.38 & 27.11 & 31.08 & 44.08 & 3.76 & 13.55 & 25.99 & 25.30 & 33.15 \\
\midrule
{\cellcolor[rgb]{0.925,0.957,1}}\textbf{GMSA} & {\cellcolor[rgb]{0.925,0.957,1}}\textbf{56.87} & {\cellcolor[rgb]{0.925,0.957,1}}\textbf{55.49} & {\cellcolor[rgb]{0.925,0.957,1}}\textbf{46.75} & {\cellcolor[rgb]{0.925,0.957,1}}\textbf{53.54} & {\cellcolor[rgb]{0.925,0.957,1}}\textbf{44.26} & {\cellcolor[rgb]{0.925,0.957,1}}\textbf{57.85} & {\cellcolor[rgb]{0.925,0.957,1}}9.40 & {\cellcolor[rgb]{0.925,0.957,1}}\textbf{19.19} & {\cellcolor[rgb]{0.925,0.957,1}}\textbf{35.36} & {\cellcolor[rgb]{0.925,0.957,1}}\textbf{39.32} & {\cellcolor[rgb]{0.925,0.957,1}}\textbf{44.29} \\
\midrule
\multicolumn{12}{c}{\textit{8x Compression Constraint}} \\
\midrule
ICAE & 25.46 & 27.83 & 25.61 & 29.39 & 17.60 & 25.38 & 1.97 & 10.42 & 18.96 & 17.66 & 22.40 \\
LLMLingua-2-large & 17.36 & 27.14 & 10.19 & 14.03 & 18.13 & 27.01 & 6.39 & 12.99 & 25.69 & 13.02 & 21.37 \\
Activation Beacon & 29.57 & 42.59 & 33.44 & 40.61 & 34.74 & 47.52 & 6.19 & 16.68 & 25.48 & 25.99 & 34.58 \\
Provence & 31.00 & 40.99 & 27.63 & 34.42 & 36.54 & 48.33 & 6.97 & 16.53 & 24.54 & 25.54 & 32.96 \\
EXIT & 44.67 & 51.21 & 19.78 & 25.10 & 37.67 & 52.46 & 4.98 & 14.00 & 30.62 & 26.78 & 34.68 \\
LongLLMLingua & 35.51 & 47.14 & 17.05 & 21.74 & 25.77 & 37.91 & 2.35 & 10.83 & 22.65 & 20.17 & 28.05 \\
\midrule
{\cellcolor[rgb]{0.925,0.957,1}}\textbf{GMSA} & {\cellcolor[rgb]{0.925,0.957,1}}\textbf{53.18} & {\cellcolor[rgb]{0.925,0.957,1}}\textbf{52.59} & {\cellcolor[rgb]{0.925,0.957,1}}\textbf{45.16} & {\cellcolor[rgb]{0.925,0.957,1}}\textbf{52.42} & {\cellcolor[rgb]{0.925,0.957,1}}\textbf{39.70} & {\cellcolor[rgb]{0.925,0.957,1}}\textbf{53.61} & {\cellcolor[rgb]{0.925,0.957,1}}\textbf{7.33} & {\cellcolor[rgb]{0.925,0.957,1}}\textbf{17.34} & {\cellcolor[rgb]{0.925,0.957,1}}\textbf{33.96} & {\cellcolor[rgb]{0.925,0.957,1}}\textbf{36.34} & {\cellcolor[rgb]{0.925,0.957,1}}\textbf{41.98} \\
\midrule
\multicolumn{12}{c}{\textbf{Qwen3-4B-Instruct}} \\
\midrule
Closed-book & 10.49 & 17.44 & 13.50 & 23.36 & 12.33 & 20.03 & 0.61 & 10.17 & - & 9.23 & 17.75 \\
Original Prompt & 32.79 & 44.02 & 33.10 & 43.48 & 44.30 & 60.31 & 8.92 & 20.41 & 31.42 & 29.78 & 39.93 \\
\midrule
\multicolumn{12}{c}{\textit{4x Compression Constraint}} \\
\midrule
ICAE & 18.64 & 20.64 & 25.24 & 29.18 & 17.77 & 25.77 & 3.03 & 11.13 & 23.13 & 16.17 & 21.97 \\
LLMLingua-2-large & 23.22 & 35.25 & 25.58 & 31.15 & 28.20 & 40.96 & 7.47 & 18.37 & 29.36 & 21.12 & 31.02 \\
Provence & 31.18 & 43.89 & 39.51 & 48.61 & 42.10 & 56.15 & 11.35 & \textbf{24.45} & 28.82 & 31.04 & 40.38 \\
EXIT & 40.00 & 52.28 & 36.97 & 45.62 & 45.16 & 57.47 & 4.99 & 15.61 & 32.03 & 31.78 & 40.60 \\
LongLLMLingua & 40.11 & 53.09 & 26.92 & 31.93 & 30.28 & 43.36 & 3.29 & 11.88 & 24.27 & 25.15 & 32.91 \\
\midrule
{\cellcolor[rgb]{0.925,0.957,1}}\textbf{GMSA} & {\cellcolor[rgb]{0.925,0.957,1}}\textbf{60.38} & {\cellcolor[rgb]{0.925,0.957,1}}\textbf{58.09} & {\cellcolor[rgb]{0.925,0.957,1}}\textbf{55.75} & {\cellcolor[rgb]{0.925,0.957,1}}\textbf{63.07} & {\cellcolor[rgb]{0.925,0.957,1}}\textbf{52.28} & {\cellcolor[rgb]{0.925,0.957,1}}\textbf{67.93} & {\cellcolor[rgb]{0.925,0.957,1}}\textbf{12.97} & {\cellcolor[rgb]{0.925,0.957,1}}24.12 & {\cellcolor[rgb]{0.925,0.957,1}}\textbf{36.26} & {\cellcolor[rgb]{0.925,0.957,1}}\textbf{45.35} & {\cellcolor[rgb]{0.925,0.957,1}}\textbf{49.89} \\
\midrule
\multicolumn{12}{c}{\textit{8x Compression Constraint}} \\
\midrule
ICAE & 18.61 & 21.41 & 24.68 & 29.69 & 17.61 & 25.69 & 2.67 & 11.82 & 23.38 & 15.89 & 22.40 \\
LLMLingua-2-large & 14.73 & 26.58 & 21.54 & 25.89 & 19.15 & 28.02 & 5.76 & 15.57 & 26.71 & 15.30 & 24.55 \\
Provence & 31.12 & 42.52 & 37.57 & 45.05 & 37.32 & 48.46 & 9.95 & 21.05 & 24.81 & 28.99 & 36.38 \\
EXIT & 41.45 & 53.38 & 29.63 & 35.01 & 35.25 & 47.83 & 2.16 & 11.55 & 28.26 & 27.12 & 35.21 \\
LongLLMLingua & 31.98 & 45.16 & 23.07 & 27.62 & 24.00 & 35.76 & 1.48 & 9.87 & 19.84 & 20.13 & 27.65 \\
\midrule
{\cellcolor[rgb]{0.925,0.957,1}}\textbf{GMSA} & {\cellcolor[rgb]{0.925,0.957,1}}\textbf{50.02} & {\cellcolor[rgb]{0.925,0.957,1}}\textbf{51.43} & {\cellcolor[rgb]{0.925,0.957,1}}\textbf{50.30} & {\cellcolor[rgb]{0.925,0.957,1}}\textbf{57.83} & {\cellcolor[rgb]{0.925,0.957,1}}\textbf{43.93} & {\cellcolor[rgb]{0.925,0.957,1}}\textbf{59.48} & {\cellcolor[rgb]{0.925,0.957,1}}\textbf{10.24} & {\cellcolor[rgb]{0.925,0.957,1}}\textbf{21.18} & {\cellcolor[rgb]{0.925,0.957,1}}\textbf{33.94} & {\cellcolor[rgb]{0.925,0.957,1}}\textbf{38.62} & {\cellcolor[rgb]{0.925,0.957,1}}\textbf{44.77} \\
\bottomrule
\end{tabular}
\end{table*}

We analyze the performance of GMSA along two core dimensions: context reconstruction capability and downstream task effectiveness under compression.

\paragraph{RQ1: Performance on context reconstruction Task.}
As shown in Figure~\ref{fig:restatement_comparasion}, GMSA-AE consistently outperforms ICAE-AE across all evaluation metrics on the context reconstruction task using the PwC dataset. Specifically, under both 4x and 8x compression ratios, GMSA-AE achieves significantly higher scores in token-matching metrics such as BLEU, Prefix Exact Match, and ROUGE-1/2/L. For instance, at sequence length 512, GMSA-AE’s BLEU score exceeds that of ICAE-AE by approximately 20--30\%, demonstrating its superior ability to precisely recover individual tokens. Furthermore, GMSA-AE also maintains a consistent 5\% advantage in BERT Score F1, which measures semantic similarity, indicating that it better preserves overall contextual semantics during compression. This confirms GMSA’s effectiveness in encoding and reconstructing semantic information even under various compression.
\paragraph{RQ2: Effectiveness on Downstream QA and Summarization Tasks under Compression.}
Table~\ref{tab:overall_comparison} presents results on multiple long-context QA and summarization benchmarks. GMSA superior performance under both 4x and 8x compression constraints across diverse models (LLaMA-3.1-8B-Instruct and Qwen3-4B-Instruct). Notably, despite employing a \textit{query-independent} compression mechanism, GMSA consistently surpasses query-dependent methods like LongLLMLingua, Provence and EXIT. This highlights GMSA’s ability to extract knowledge without relying on query-specific signals. Moreover, GMSA demonstrates more robust and stable gains across varying compression constraint and model scales, further validating the generalizability and effectiveness of GMSA.

\subsection{Efficiency Analysis}
In this section, we discuss the efficiency of our proposed method. By using soft tokens instead of the long original context to enhance the inference process, our method reduces the inference cost of the original context during the generation process by a factor of $r$. The overall floating-point operations (FLOPs) are calculated through two processes: compression and generation.

\begin{figure*}[htb]
\centering
\begin{subfigure}[b]{0.32\linewidth}
  \centering
  \includegraphics[width=\linewidth]{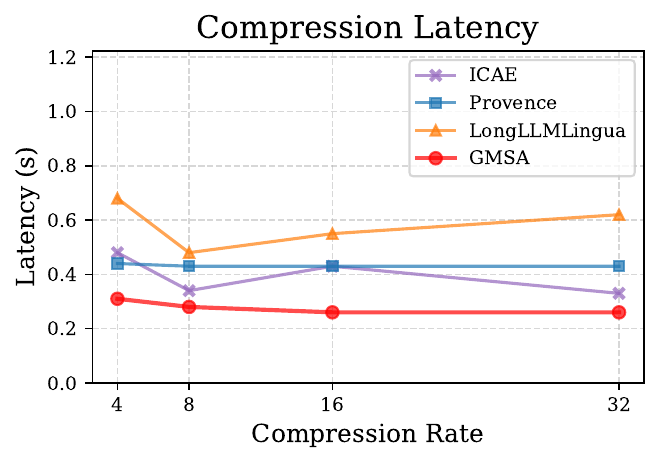}
  \caption{Compression Latency.}
  \label{fig:1a}
\end{subfigure}
\hfill
\begin{subfigure}[b]{0.32\linewidth}
  \centering
  \includegraphics[width=\linewidth]{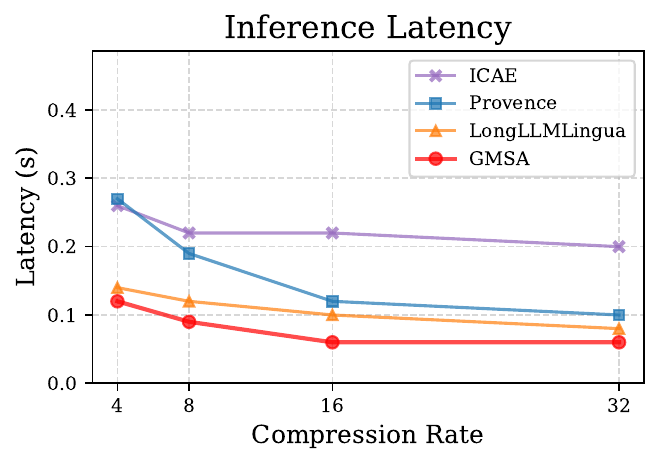}
  \caption{Inference Latency.}
  \label{fig:1b}
\end{subfigure}
\hfill
\begin{subfigure}[b]{0.32\linewidth}
  \centering
  \includegraphics[width=\linewidth]{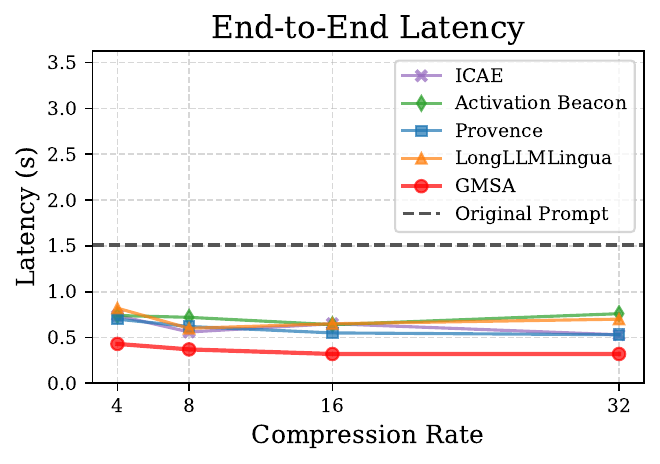}
  \caption{End-to-End Latency.}
  \label{fig:1c}
\end{subfigure}
\caption{Latency evaluation across methods. For context compression methods, end-to-end inference latency can be decomposed into compression latency and inference latency. Activation Beacon does not explicitly decouple these two phases and therefore reports only the end-to-end latency. Detailed numerical results are provided in Appendix~\ref{apx:numerical_result}.}
\label{fig:lat_eval}
\end{figure*}

The compression process can be expressed as:
$$
\mathrm{FLOPs}^{comp} = F^{\mathrm{Encoder}}(L) + F^{\mathrm{LSA}}\left(\left\lceil \frac{L}{r} \right\rceil\right) \, .
$$
were $L$ denotes the original context length; $L_q$ refers to the question length, and $F^{*}(\cdot)$ represents the FLOPs complexity measure for module $*$. The symbol $*$ indicates the architectural components, where $* \in \{\mathrm{Decoder}, \mathrm{Encoder}, \mathrm{LSA}\}$.
For the generation process, assuming the answer length is $L_a$, the generation process requires $L_a$ forward passes. The FLOPs for the $i$-th forward pass are given by:
$$
\mathrm{FLOPs}^{forward}_{i} = F^{\mathrm{Decoder}}\left(\left\lceil \frac{L}{r} \right\rceil, L_q, i\right) \, .
$$
Combining the costs of all components, the total FLOPs complexity is:
$$
\mathrm{FLOPs} = \sum_{i=1}^{L_a} \mathrm{FLOPs}_i^{\text{forward}} + \mathrm{FLOPs}^{comp} \, .
$$
GMSA achieves the lowest end-to-end inference latency against, which is approximately 5x faster than original prompt input under 32x compression rate on NaturalQuestions (see Figure~\ref{fig:lat_eval} and Appendix~\ref{apx:numerical_result}).

\begin{table}[htb]
\centering
\small
\caption{Ablation study on NaturalQuestions, 2WikiMQA under 4x compression constraint using Qwen3-4B as backbone.}
\begin{tabular}{ l | c c c c}
\toprule
\multirow{2}{*}{\textbf{Methods}} & 
  \multicolumn{2}{c}{\textbf{NaturalQA}} & 
  \multicolumn{2}{c}{\textbf{2WikiMQA}} \\
\cmidrule(lr){2-3} \cmidrule(lr){4-5}
& \textbf{EM} & \textbf{F1} & \textbf{EM} & \textbf{F1} \\
\midrule
{\cellcolor[rgb]{0.925,0.957,1}}\textbf{Default} & {\cellcolor[rgb]{0.925,0.957,1}}\textbf{60.38} & {\cellcolor[rgb]{0.925,0.957,1}}\textbf{58.09} & {\cellcolor[rgb]{0.925,0.957,1}}\textbf{55.75} & {\cellcolor[rgb]{0.925,0.957,1}}\textbf{63.07} \\
\midrule
\emph{w/o} AE Training & 48.81 & 51.65 & 37.51 & 42.01 \\
\emph{w/o} Group Merging & 21.88 & 24.55 & 28.33 & 32.02 \\
\emph{w/o} LSA & 20.15 & 22.30 & 27.74 & 31.21 \\
\emph{w/} Rand. init. LSA & 19.47 & 21.99 & 28.75 & 32.36 \\
\bottomrule
\end{tabular}
\label{tab:ablation}
\end{table}

\begin{figure}[htb]
    \centering
    \includegraphics[width=0.8\linewidth]{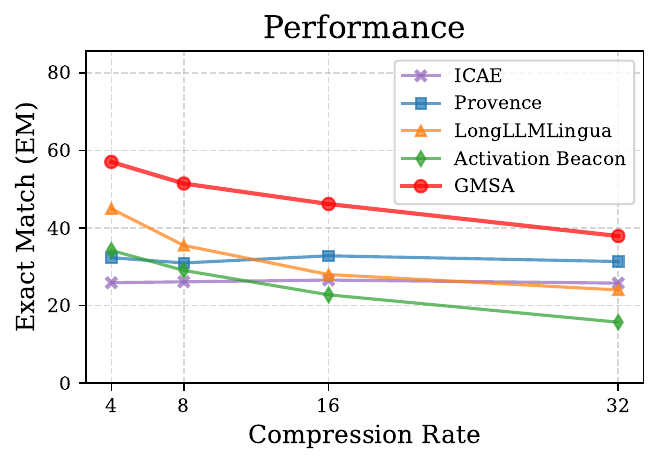}
    \caption{Performance comparison and stress test of GMSA and baselines across various compression rate on the NaturalQuestions dataset, using LLaMA-3.2-3B as the backbone.}
    \label{fig:stress_test}
\end{figure}

\subsection{Ablation Study}
For RQ3, to investigate the impact of each component in GMSA, we conduct the following four ablation experiments (see Table~\ref{tab:ablation}): (1) Ours \textit{w/o} AE Training refers to directly perform fine-tuning on GMSA without AutoEncoding (AE) training; (2) Ours \textit{w/o} Group Merging indicates that we replace group merging with appending learnable tokens when generating summary vectors; (3) Ours \textit{w/o} LSA means we do not use the Layer Semantic Alignment (LSA) module, but instead directly pass the summary vectors through an MultiLayer Perceptron (MLP) to obtain the summary vectors; (4) Ours \textit{w/} Rand.~init.~LSA denotes that the LSA module is randomly initialized instead of inheriting the pre-trained weights from the decoder.


In summary, removing any single component leads to a significant performance drop, fully demonstrating the necessity and effectiveness of each component. Removing autoencoder pretraining makes it difficult for GMSA to generate summary vectors that capture complete semantics; replacing Group Merging with learnable appended tokens increases the model's learning difficulty; discarding the Layer Semantic Alignment (LSA) module results in misalignment between the high-level semantic information represented by the summary vectors and the low-level semantic space of the decoder’s input; and randomly initializing LSA fundamentally prevents it from inheriting the decoder’s initial input semantic space, thereby introducing a substantial semantic gap that is difficult to bridge even through training. Further evidence is provided in Appendix~\ref{sec:loss_compara}: under all ablation settings, loss convergence is significantly worse than that of the Default setting, both during the autoencoder pretraining phase and the fine-tuning phase.

\subsection{Stress Test}
As shown in Figure~\ref{fig:stress_test}, we conduct stress tests on GMSA and its baseline methods across multiple compression rates (2x, 4x, 8x, 16x, and 32x) on the NaturalQuestions dataset. Experimental results show that although certain methods (e.g., LongLLMLingua and Activation Beacon) suffer significant performance degradation under high compression, GMSA consistently maintains the best performance across all compression ratios, with a smaller performance drop compared to most baselines. Notably, while methods such as Providence and ICAE exhibit relatively stable performance across different compression levels, their overall accuracy remains substantially lower than that of GMSA.

\section{Conclusion}

This paper introduces GMSA, a context compression framework based on an encoder-decoder structure. It effectively and efficiently learns summary vectors and bridges the significant gap between the semantics representation of different layers via group merging, and a Layer Semantic Alignment (LSA) module. GMSA first undergoes autoencoder pretraining to ensure that the generated soft tokens contain complete semantics, and then adapts to downstream tasks through fine-tuning. Experiments demonstrate that GMSA has excellent context reconstruction capabilities. It outperforms existing baselines by a large margin in downstream tasks, paving the way for the efficient application of LLMs.

\section*{Limitations}
Although GMSA demonstrates strong performance across diverse long-context benchmarks and compression ratios, it has a limitation. Like most soft prompt compression methods, GMSA currently requires a two-stage training pipeline, consisting of autoencoder pretraining followed by task-specific fine-tuning, which incurs modest overhead compared to training-free compression baselines. However, this design is difficult to avoid in task-agnostic compression frameworks, and GMSA’s superior downstream performance and inference efficiency adequately compensate for this overhead.

\bibliography{custom}

\clearpage
\appendix

\section{Impact of different number of LSA layers}
\label{apx:lsa_lays}
\begin{figure*}[htb]
    \centering
    \includegraphics[width=0.8\linewidth]{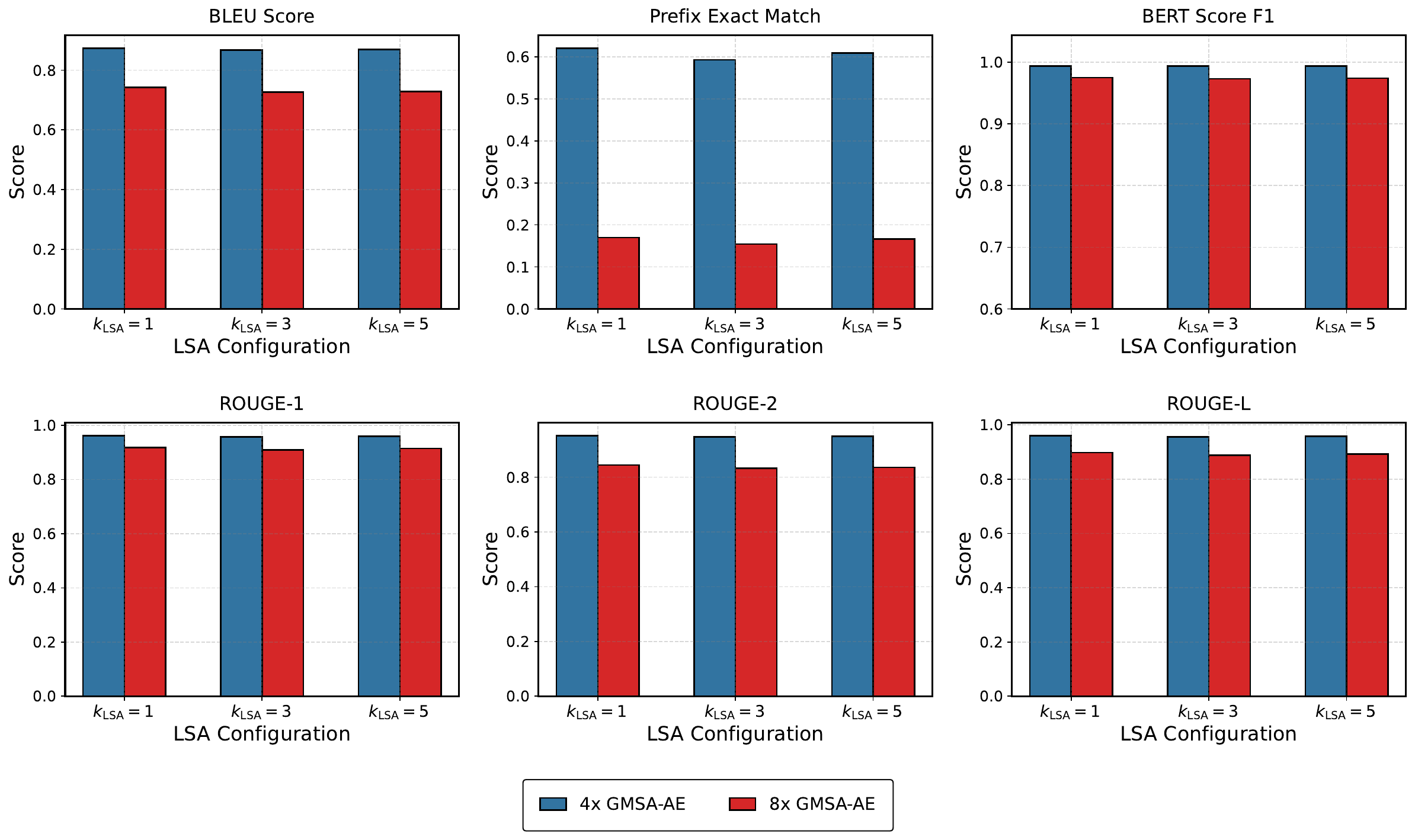}
    \caption{The impact of different number of layers of LSA on semantic retention in GMSA-AE. Sequence Length is set to 512}
    \label{fig:lsa_comparasion}
\end{figure*}

We conduct experiments to investigate the impact of Layer Semantic Alignment (LSA) module with varying numbers of layers on the retention of complete semantics, and the results are shown in Figure~\ref{fig:lsa_comparasion}. We can draw the following conclusions: (1) Only one layer of LSA is sufficient to achieve good retention of complete semantics (with a BERT Score F1 close to 1, and it already performs the best among different numbers of LSA layers); (2) When the number of LSA layers becomes too high, e.g., using five layers of LSA, it may actually lead to a decrease in the GMSA's ability to retain semantics. This is likely because as the LSA module becomes deeper, it contains more high-layer semantics and fewer low-layer semantics, thereby increasing the difficulty of semantic alignment.

\begin{table}[htbp]
\small
\centering
\caption{Latency evaluation on NaturalQA using LLaMA-3.2-3B as backbone.  
Each compression method's total latency can be divided into compression latency and inference latency. }
\label{tab:latency_results}
\begin{tabular}{l|cccc}
\toprule
\textbf{Methods} & \multicolumn{4}{c}{\textbf{Compression Constraint}} \\
\cmidrule(lr){2-5}
& \textbf{4x} & \textbf{8x} & \textbf{16x} & \textbf{32x} \\
\midrule
\multicolumn{5}{c}{\textbf{Compression Latency (s)}} \\
\midrule
ICAE & 0.48 & 0.34 & 0.43 & 0.33 \\
Provence & 0.44 & 0.43 & 0.43 & 0.43 \\
LongLLMLingua & 0.68 & 0.48 & 0.55 & 0.62 \\
\midrule
{\cellcolor[rgb]{0.925,0.957,1}}\textbf{GMSA} & {\cellcolor[rgb]{0.925,0.957,1}}\textbf{0.31} & {\cellcolor[rgb]{0.925,0.957,1}}\textbf{0.28} & {\cellcolor[rgb]{0.925,0.957,1}}\textbf{0.26} & {\cellcolor[rgb]{0.925,0.957,1}}\textbf{0.26} \\
\midrule
\multicolumn{5}{c}{\textbf{Inference Latency (s)}} \\
\midrule
ICAE & 0.26 & 0.22 & 0.22 & 0.20 \\
Provence & 0.27 & 0.19 & 0.12 & 0.10 \\
LongLLMLingua & 0.14 & 0.12 & 0.10 & 0.08 \\
\midrule
{\cellcolor[rgb]{0.925,0.957,1}}\textbf{GMSA} & {\cellcolor[rgb]{0.925,0.957,1}}\textbf{0.12} & {\cellcolor[rgb]{0.925,0.957,1}}\textbf{0.09} & {\cellcolor[rgb]{0.925,0.957,1}}\textbf{0.06} & {\cellcolor[rgb]{0.925,0.957,1}}\textbf{0.06} \\
\midrule
\multicolumn{5}{c}{\textbf{End-to-End Latency (s)}} \\
\midrule
ICAE & 0.74 & 0.56 & 0.65 & 0.53 \\
Activation Beacon & 0.74 & 0.72 & 0.64 & 0.76 \\
Provence & 2.54 & 2.43 & 0.55 & 0.53 \\
LongLLMLingua & 0.82 & 0.60 & 0.65 & 0.70 \\
\midrule
{\cellcolor[rgb]{0.925,0.957,1}}\textbf{GMSA} & {\cellcolor[rgb]{0.925,0.957,1}}\textbf{0.43} & {\cellcolor[rgb]{0.925,0.957,1}}\textbf{0.37} & {\cellcolor[rgb]{0.925,0.957,1}}\textbf{0.32} & {\cellcolor[rgb]{0.925,0.957,1}}\textbf{0.32} \\
\midrule
Original Prompt & \multicolumn{4}{c}{1.51}\\
\bottomrule
\end{tabular}
\end{table}

\section{Training Loss Comparison}
\label{sec:loss_compara}
\begin{figure*}[htb]
    \centering
    \begin{subfigure}[b]{0.8\linewidth}
        \centering
        \includegraphics[width=\linewidth]{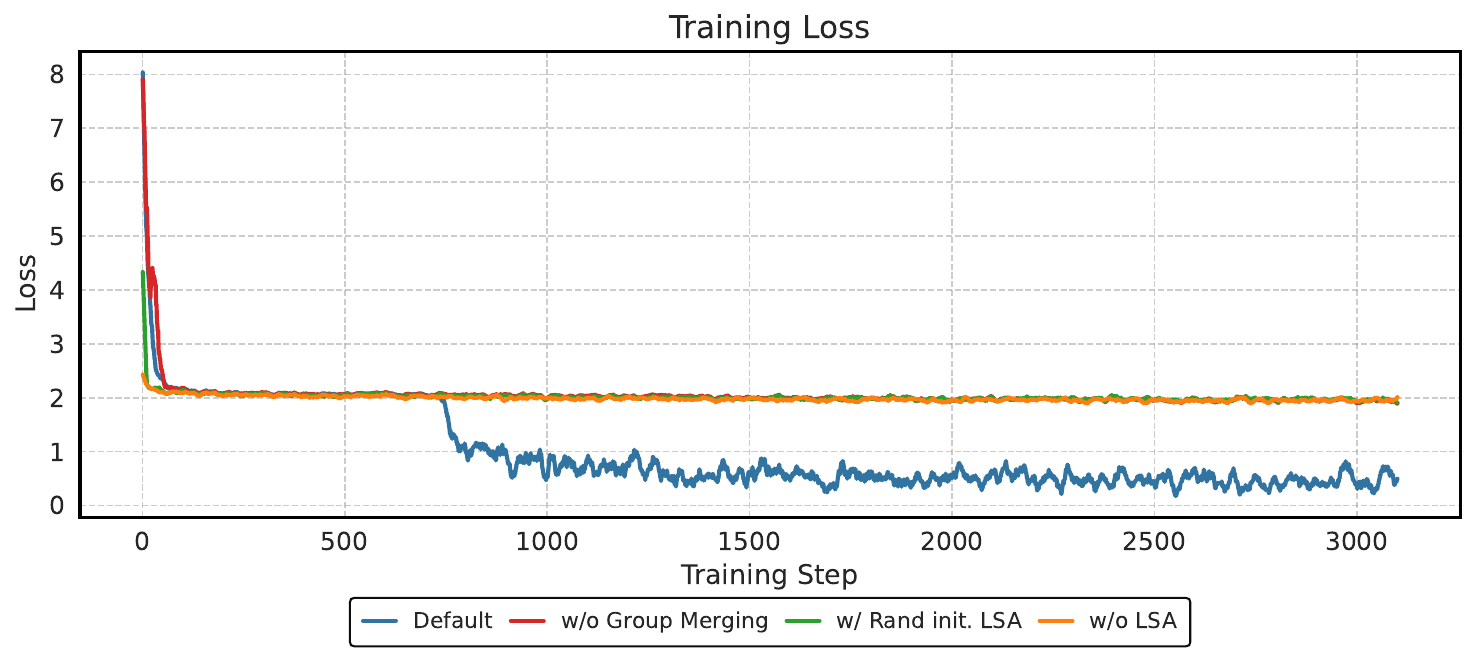}
        \caption{Loss comparison during the autoencoder pretraining stage.}
        \label{fig:ae_loss_compara}
    \end{subfigure}
    
    \vspace{1em} 
    
    \begin{subfigure}[b]{0.8\linewidth}
        \centering
        \includegraphics[width=\linewidth]{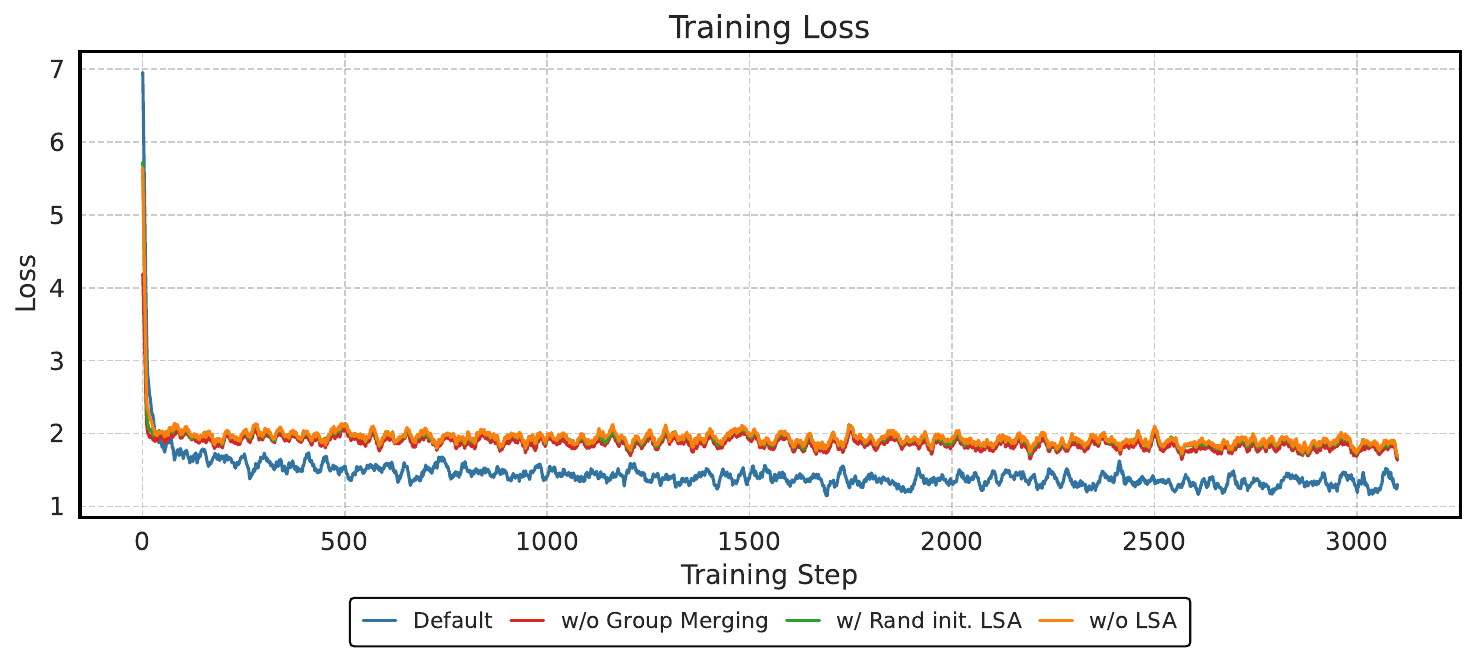}
        \caption{Loss comparison during the fine-tuning stage.}
        \label{fig:ft_loss_compara}
    \end{subfigure}
    
    \caption{
        Loss comparison of two training stage. (a) and (b) show loss comparisons during autoencoder pretraining and fine-tuning stages, respectively. 
        \textbf{Note: \emph{w/o} Group Merging, \emph{w/} Rand init. LSA, and \emph{w/o} LSA show nearly identical loss curves in both stages.}
    }
    \label{fig:loss_comparison}
\end{figure*}

Figure~\ref{fig:ae_loss_compara} shows the loss curves during the autoencoder pretraining phase under four settings: Default, \emph{w/o} Group Merging, \emph{w/} Rand Init. LSA, and \emph{w/o} LSA. The Default setting converges substantially faster and achieves a lower final loss, whereas the other three settings exhibit similar loss trajectories with significantly higher losses. This indicates that, compared to the Default configuration, each of these ablated variants incurs considerable semantic degradation.

During the fine-tuning phase (Figure~\ref{fig:ft_loss_compara}), the training loss of the three configurations (i.e., \textit{w/o} Group Merging, \textit{w/} Rand init.\ LSA, and \textit{w/o} LSA) are significantly higher than that of the \textit{Default} setting, with similar loss trajectories. This suggests that these configurations are less effective at extracting knowledge from the compressed representation compared to the default setup.



\section{Datasets Details}
\label{apx:data_details}
\paragraph{PwC dataset.} In the PwC dataset~\cite{ge2024incontext}, each sample is a triplet (context, prompt, answer), where the context is sampled from the Pile and the prompt and answer are generated by GPT-4. The training set contains 241,564 samples, the test set contains 18,146 samples.

\paragraph{NaturalQuestions.} 
NaturalQuestions~\cite{liu-etal-2024-lost}, in which each question corresponds to 20 relevant documents, 19 of which are distractors and only one contains the ground truth answer. The training set contains 75,322 samples, the test set contains 2,655 samples.

\paragraph{HotpotQA.}
HotpotQA~\cite{yang-etal-2018-hotpotqa} is a two-hop reasoning dataset, where the answers are scattered across two documents. Specifically, each question corresponds to 10 relevant documents, two of which are the ground truth documents. The training set contains 89,609 samples, the test set contains 7,345 samples.

\paragraph{2WikiMQA.}
Compared with HotpotQA, 2WikiMQA~\cite{ho-etal-2020-constructing} includes more complex reasoning paths, and the combination of structured and unstructured data, usually involving two or more hops and having higher difficulty. The training set contains 167,454 samples, the test set contains 12,576 samples.

\paragraph{NarrativeQA.}
NarrativeQA~\cite{DBLP:journals/tacl/KociskySBDHMG18} is a reading comprehension dataset designed to evaluate deeper narrative understanding, where models must answer questions about stories by reading entire books or movie scripts rather than relying on shallow lexical matching. The dataset includes document-level metadata (e.g., story URLs and approximate word counts), Wikipedia summaries, and question-answer pairs. We use the test set to evaluate the model's performance, filtering out test samples longer than 32K.

\paragraph{MultiNews.}
MultiNews~\cite{DBLP:conf/acl/FabbriLSLR19} is a multi-document summarization dataset, where each sample consists of multiple news articles paired with a human-written summary. The dataset provides 44,972 training examples, 5,622 validation examples, and 5,622 test examples. We use the test set to evaluate model performance.

\section{Numerical Latency Evaluation Results}
\label{apx:numerical_result}
Table~\ref{tab:latency_results} presents the numerical results corresponding to Figure~\ref{fig:lat_eval}. We can see that GMSA consistently achieves the lowest compression latency, inference latency, and end-to-end latency, and is significantly faster than the original prompt, e.g., up to 8× faster under an 8x compression constraint.

\section{Ethic Statement}
This paper introduces GMSA, a context compression framework based on the encoder-decoder architecture. It effectively and efficiently learns summary vectors and bridges the significant gap between  different layers via group merging, and a LSA module. The data and models used in our research are released under open-source licenses and sourced from open platforms. Although our work may have various societal impacts, it does not introduce any additional ethical concerns compared to existing text compression methods. Therefore, we believe it is unnecessary to specifically highlight any particular ethical issues here.

\section{Language Model Usage Statement}
A large language model was used during manuscript preparation solely for editorial support. Its contributions were limited to language polishing and clarity enhancement (e.g., rephrasing, tightening explanations, and improving the presentation of formulas/derivations), and it did not generate the underlying research content. All central concepts, methodological decisions, experimental procedures, and empirical findings are the authors' original work.

\end{document}